%% file: main.tex
\documentclass[11pt]{article}
\PassOptionsToPackage{table}{xcolor}
\PassOptionsToPackage{hypertexnames=false}{hyperref}

\usepackage[letterpaper]{naiss-hku-template} 

\usepackage{url}
\usepackage{xurl} 
\usepackage{wrapfig} 
\usepackage{subcaption} 
\usepackage{booktabs}
\usepackage{multirow}
\usepackage{bbm} 
\usepackage{algorithm}
\usepackage{algpseudocode}

\input{math_commands.tex}

\definecolor{baserow}{gray}{0.85}
\definecolor{lightgreen}{rgb}{0.9,1.0,0.9}
\colorlet{oursrow}{lightgreen}

\graphicspath{{figures/}}

\setNAISSTitle{Balancing Capability Acquisition and Retention in Offline Agent Fine-Tuning}

\setNAISSAuthors{%
  Ronghua Li\textsuperscript{1} \quad
  Zi Liang\textsuperscript{2} \quad
  Zhishan Li\textsuperscript{3} \quad
  Shinan Liu\textsuperscript{1}}
\setNAISSAffiliation{%
  \textsuperscript{1}The University of Hong Kong \qquad
  \textsuperscript{2}The Hong Kong Polytechnic University\\[0.15em]
  \textsuperscript{3}Independent Researcher}
\setNAISSEmail{ \{coryli,shinan6\}@hku.hk, zi1415926.liang@connect.polyu.hk, zhishanli@zju.edu.cn}
\setNAISSAbstract{%
Supervised fine-tuning (SFT) on offline agent trajectories is the standard approach for training specialized tool-using agents, but forcing models to imitate reasoning and actions token by token may harm other capabilities (e.g., general reasoning, tool calling, code generation) of the base model. In this work, we focus on studying \emph{how to better balance the trade-off between acquiring new capabilities and preserving existing ones during agent trace SFT}. By comparing several baselines in our setup, standard SFT improves the target benchmark while lowering several non-target benchmark scores; meanwhile, simply constraining distributional drift using KL penalty or limiting the update magnitude did not avoid this regression trend. Motivated by recent token-wise adaptive learning objectives, this work proposes \textbf{Privilege-Guided SFT (PG-SFT)} to leverage turn-level information gain of agent trajectories as an indicator to adjust supervision strength. PG-SFT yields a more favorable observed trade-off on the evaluated benchmarks, substantially reducing distributional drift and broad capability degradation at the cost of slight degradation in target-task performance. Our findings suggest that balancing the acquisition--retention trade-off depends not only on whether the model is anchored to its base behavior, but also on where and how strongly supervision should depart from that behavior.}

\setNAISSLinks{%
  \href{https://github.com/66Ronghua99/PGSFT}{%
    \tikz[baseline=(code.base)]{%
      \node[fill=NAISSGreen, text=white, rounded corners=3pt,
            inner xsep=7pt, inner ysep=3.5pt] (code) {\faGithub{} Code};}%
  }%
}

\begin{document}
\NAISSMakeTitle
\NAISSMakeAbstract

\input{sections/introduction}
\input{sections/related_work}
\input{sections/method}
\input{sections/experiments}
\input{sections/conclusion}


\bibliographystyle{plainnat}
\bibliography{references}

\appendix
\input{sections/appendix}

\end{document}

%% file: math_commands.tex
\usepackage{amsmath,amsfonts,bm}

\def\eqref#1{equation~\ref{#1}}

\def\1{\bm{1}}

\DeclareMathAlphabet{\mathsfit}{\encodingdefault}{\sfdefault}{m}{sl}
\SetMathAlphabet{\mathsfit}{bold}{\encodingdefault}{\sfdefault}{bx}{n}



%% file: sections/introduction.tex
\section{Introduction}\label{sec:introduction}

Supervised fine-tuning (SFT) has been the standard approach for adapting large language models (LLMs) to follow instructions and perform specialized tasks \citep{ouyang2022instructgpt,touvron2023llama2}. For tool-using agents, offline execution trajectories provide supervision on reasoning and actions, allowing models to acquire task-specific behaviors without repeatedly interacting with expensive environments during training~\citep{yao2023react,chen2024agentflan,zeng2023agenttuning}. 

However, token-level imitation treats every demonstrated assistant token as a target, regardless of whether it represents a task-relevant decision or an incidental realization of the trajectory. Fine-tuning can thus improve the target workflow while changing capabilities acquired before adaptation, a tension related to catastrophic forgetting~\citep{kirkpatrick2017overcoming,li2017learning,kemker2018measuring}. Regressions in knowledge, reasoning, instruction following, and safety-related behavior have also been observed in LLM post-training~\citep{luo2023forgetting,qi2023safety,wang2026memsft,xu2026masktarget,lin2025sft,huan2025math,wang2024promot,biderman2024lora}. Prior approaches reduce model--data mismatch through self-distillation~\citep{yang2024sdft} or constrain changes from the base model through reference-based objectives~\citep{zhu2025asft,xu2026masktarget}, but whether such control suffices to preserve capabilities remains undemonstrated; moreover, they focus on reasoning and instruction-following tasks, leaving capability retention in agent fine-tuning largely unexplored.

In this work, we study the following question: \textbf{How does supervision-objective design affect the empirical trade-off between target adaptation and non-target benchmark performance?}

Intuitively, SFT may disrupt existing capabilities less if the student stays close to its base distribution---by anchoring predictions to it or limiting successive update sizes. Yet, a model that changes too little may retain its old behavior without acquiring the target behavior. We examine this tension by holding the offline trajectories fixed and varying only the supervision objective: standard SFT provides the full-imitation reference; SFT with a Base-model KL penalty tests direct anchoring to the base distribution; and proximal SFT (PSFT) tests a constraint on updates relative to a lagged policy~\citep{zhu2026psft}. Together, these methods ask whether conservative learning suffices for balancing capability acquisition and retention.

Unfortunately, extensive experiments on Qwen3~\citep{qwen2025qwen3} and Qwen3.5~\citep{qwen2026qwen35} across four benchmarks show that better fitting of the training data can coincide with larger distributional deviation from the base model, yet simply reducing this deviation with a fixed KL constraint does not necessarily produce better downstream capability retention. Trust-region baseline PSFT keeps the model closest to the base distribution, but acquires the least of the target behavior. 

Recently, token-wise supervision adjustment DFT~\citep{wu2025dft} and InfoSFT~\citep{sabbaghi2026infosft} yielded good performance in reasoning and math benchmarks. Although our preliminary analyses reveal DFT's incompatibility with agent SFT, their insights on adaptive supervision motivate us to introduce \textbf{Privilege-Guided Supervised Fine-Tuning (PG-SFT)}. PG-SFT allocates supervision turn by turn, matching the granularity at which agent decisions appear, and uses ``next actions'' in trajectories as training-time privileged information to measure the ``knowledge gain'' imposed by the hint, adjusting the supervision intensity of that turn accordingly. On Qwen3.5-4B, PG-SFT matches SFT on the selected target SWE-bench subset (59/89, pass@3) while retaining stronger performance on other non-target benchmarks. On Qwen3-4B-Thinking, PG-SFT gives up limited target-task performance in exchange for stronger retention on non-target tasks. These results suggest that the key of balancing the trade-off is not only regularizing supervision toward the base model, but deciding \emph{where and how strongly} the model should depart from it.

Our contributions are summarized as follows:
\begin{itemize}
    \item \textbf{An empirical study of capability acquisition and retention in offline agent fine-tuning.}
    We formulate post-training as a joint learning--retention problem and evaluate adaptation not only through target-task performance, but also through target fitting, distributional drift from the base model, and retained non-target capabilities.

    \item \textbf{PG-SFT as an adaptive probe of the learning--retention trade-off.}
    We introduce Privilege-Guided SFT, which uses privileged conditioning to estimate the intensity of moving away from the base distribution. PG-SFT provides a concrete mechanism for testing whether adaptive, example-dependent learning intensity can achieve a more favorable balance between acquiring specialized behavior and retaining existing capabilities.

    \item \textbf{A controlled comparison of different learning pressures.}
    Using the same offline trajectories, we compare ordinary SFT, SFT with a fixed Base-KL constraint, a clipped-ratio trust-region objective PSFT, and an adaptive supervision strategy PG-SFT. The comparison shows that fitting the target distribution, remaining close to the base distribution, and preserving downstream capabilities are distinct properties; optimizing one does not automatically guarantee the others.

\end{itemize}

The remainder of this paper is organized as follows: Section~\ref{sec:related} reviews related work; Section~\ref{sec:method} presents the compared objectives and PG-SFT; Section~\ref{sec:experiments} reports the acquisition--retention analyses; Section~\ref{sec:conclusion} analyzes the turn-level coefficient and discusses the scope of our study.

%% file: sections/related_work.tex
\section{Related Work}
\label{sec:related}

\subsection{Capability retention in post-training}
Fine-tuning enables language models to acquire specialized behaviors, but adaptation can also modify capabilities that are not directly optimized \cite{lin2025sft,huan2025math,wang2024promot}: continual fine-tuning studies document forgetting of previously acquired knowledge and tasks \cite{luo2023forgetting,biderman2024lora}, and alignment-oriented investigations show that task-specific adaptation may affect safety-related behaviors \cite{qi2023safety}. Existing mitigations either constrain deviation from the initial model through reference-based objectives or update rules \cite{yang2024sdft,zhu2025asft,xu2026masktarget,zhu2026psft,li2018l2sp}, or preserve the original parameters through additional components \cite{wang2026memsft,shenfeld2025razor}.

\subsection{Model-aware supervision and adaptive learning pressure}
Several observations suggest that supervision should be calibrated to the learner: data fitting the learner is more effective \cite{zhang2025grape}, self-distillation reduces model--data mismatch by generating supervision closer to the learner distribution \cite{yang2024sdft}, and smaller models struggle to absorb reasoning behaviors from substantially stronger teachers \cite{li2025small}. Some work curates the data by it self to improve data-model alignment, including, quality filtering~\cite{chen2023alpagasus,liu2024deita}, difficulty grading~\cite{li2024ifd,liu2024selectit}, and data rewriting~\cite{li2024reflection}, though such gains do not always scale to large pools~\cite{xia2025rethinking}. Some recent studies proposed to adjust supervision strength non-uniformly to better match the data-model relationship: DFT scales token-level updates by the model's confidence on demonstrations \cite{wu2025dft}, IDFT accounts for contextual distribution differences \cite{zhang2026idft}, anchored SFT combines adaptive weighting with reference-model regularization \cite{zhu2025asft}, CADFT estimates example-level compatibility \cite{zhou2026cadft}, and TALR reweights per-token losses to reduce interference with general capabilities \cite{lin2025sft}. PSFT also adapts the contribution of supervised tokens, but its signal derives from the magnitude of previous model updates~\cite{zhu2026psft}.

\subsection{Privileged information as a training-time signal}
Learning using privileged information studies settings where additional information is available during training but unavailable during inference \cite{vapnik2015lupi}, and generalized distillation shows that such training-only signals improve prediction when they inform the target mapping \cite{lopezdpaz2016unifying}. Privileged information is not universally beneficial, however: stronger privileged features do not necessarily yield better student models, motivating selective use of privileged signals \cite{yang2022privileged,aslam2023privileged}; recent analyses further find that privileged teachers can bias solutions~\cite{harne2026privilegedbiased}, suppress self-correction~\cite{kaur2026thinking}, or yield unstable gains~\cite{shrestha2026rethinkingpi}, and that more privileged information is not automatically better~\cite{zhao2026morepi}. In language models, privileged signals have been explored through context distillation and self-distillation mechanisms \cite{snell2022distillcontext,agarwal2024gkd,zhao2026opsd,penaloza2026pidistill}.

%% file: sections/method.tex
\section{Trade-off Balancing: A Supervision Objective Perspective}
\label{sec:method}

This section investigates the acquisition--retention tension through a controlled comparison of training objectives. All objectives are trained on identical data with identical supervised tokens, so that only the objective varies. We develop the comparison along two lines. The \textbf{\emph{conservative}} line restrains the overall fitting strength with a single uniform rule, asking whether more conservative learning alone can balance acquisition against retention. The \textbf{\emph{adaptive}} line instead varies supervision strength across decisions; our first attempt in this direction, DFT~\citep{wu2025dft}, underperforms even the untrained base model. Inspired by some preliminary results, we propose PG-SFT, a turn-level adaptive objective that allocates supervision with training-time privileged information.

\subsection{Preliminaries and Metrics}
\label{sec:method_setup}

Let $p_0$ be a frozen base model, $p_\theta$ the fine-tuned student, and $\mathcal{D}$ an offline dataset of agent trajectories. User messages and tool feedback are excluded from the loss. At assistant turn $k$, $t$ is the position of the token in the turn, $y_t$ is the token, and $x_t$ is the trajectory context, including preceding turns and assistant tokens at the current turn; $V$ is the vocabulary. Under teacher forcing, we define
\begin{equation}
q_{0,t}(v)=p_0(v\mid x_t),
\qquad
p_{\theta,t}(v)=p_\theta(v\mid x_t),
\qquad v\in V.
\label{eq:ordinary_distributions}
\end{equation}

Two teacher-forced statistics recur throughout our analysis: the average negative log-likelihood (NLL) of the demonstrated tokens, which measures how strongly a model fits the demonstrations, and the average forward KL divergence from the base model (\emph{Base KL}), which measures how far its predictions move away from $p_0$ on the same contexts. Formally, for a model with per-position distributions $p_t$,
\begin{equation}
\mathrm{NLL}(p)=-\frac{1}{N}\sum_t\log p_t(y_t),
\qquad
\mathrm{KL}_{\mathrm{base}}(p)=\frac{1}{N}\sum_t
D_{\mathrm{KL}}\!\left(q_{0,t}\Vert p_t\right),
\label{eq:metrics}
\end{equation}
where $N$ is the number of supervised assistant tokens.

\subsection{Conservative Supervision Objectives}
\label{sec:learning_pressure}

\paragraph{Full fitting: SFT.}
Standard supervised fine-tuning minimizes
\begin{equation}
\mathcal{L}_{\mathrm{SFT}}(\theta)
=-\frac{1}{N}\sum_t\log p_{\theta,t}(y_t),
\label{eq:sft}
\end{equation}
whose one-hot target places all probability mass on the demonstrated token. SFT supplies the full-fitting endpoint of our comparisons.

\paragraph{Uniformly reduced fitting: Base-KL.}
A natural way to retain more of the base distribution is to add a forward KL penalty to the same supervised loss:
\begin{equation}
\mathcal{L}_{\mathrm{SFT+KL}}(\theta)
=\mathcal{L}_{\mathrm{SFT}}(\theta)
+\frac{\lambda}{N}\sum_t
D_{\mathrm{KL}}\!\left(q_{0,t}\Vert p_{\theta,t}\right),
\qquad \lambda\geq 0.
\label{eq:base_kl}
\end{equation}
Up to a positive scale and terms independent of $\theta$, this is cross-entropy against
\begin{equation}
r_t^{\mathrm{SFT+KL}}(v)
=\frac{\lambda}{1+\lambda}q_{0,t}(v)
+\frac{1}{1+\lambda}\mathbbm{1}[v=y_t].
\label{eq:fixed_target}
\end{equation}
That is, the target interpolates between the base distribution $q_{0,t}$ and the one-hot target at the demonstrated token. We name the weight on the one-hot component the supervision coefficient $\alpha$; Base-KL fixes $\alpha_{\mathrm{KL}}=1/(1+\lambda)$ for every turn. Sweeping $\lambda$ then tests how much of the acquisition--retention behavior can be explained by reducing supervision uniformly.

\paragraph{Limited updates: proximal SFT.}
A third strategy leaves the target untouched and constrains the update instead. We consider the clipped-ratio objective of \citet{zhu2026psft}, which constrains each update relative to a lagged student policy $p_{\theta_{\mathrm{old}}}$. With a constant positive advantage and clipping radius $\epsilon>0$, its surrogate is maximized as
\begin{equation}
\mathcal{J}_{\mathrm{PSFT}}(\theta)
=\mathbb{E}_{(x_t,y_t)\sim\mathcal{D}}
\left[
\min\!\left(\rho_t,
\operatorname{clip}(\rho_t,1-\epsilon,1+\epsilon)\right)
\right],
\qquad
\rho_t=\frac{p_{\theta,t}(y_t)}{p_{\theta_{\mathrm{old}},t}(y_t)}.
\label{eq:psft}
\end{equation}
This rule limits probability increases on the demonstrated token relative to the lagged reference, but defines neither an intermediate target nor an anchor of the output distribution to $q_{0,t}$; it tests whether controlling updates relative to the recent policy improves retention. We therefore compare PSFT empirically, without treating its clipping radius as an $\alpha$ value.

Together, these objectives form the conservative family of our comparison: they control how strongly the student fits the demonstrations---either by interpolating the target between the base distribution and the demonstrated token, or by limiting each update relative to the recent policy. However, none of them conditions the target on the data--model relationship. Whether \emph{adaptive} allocation of supervision matters beyond its overall strength is the question we take up next.

\subsection{Adaptive Supervision Objectives: Preliminary Attempts on Token-wise Adjustment}
\label{sec:method_motivation}

\begin{figure}[h]
    \centering
    \includegraphics[width=\textwidth]{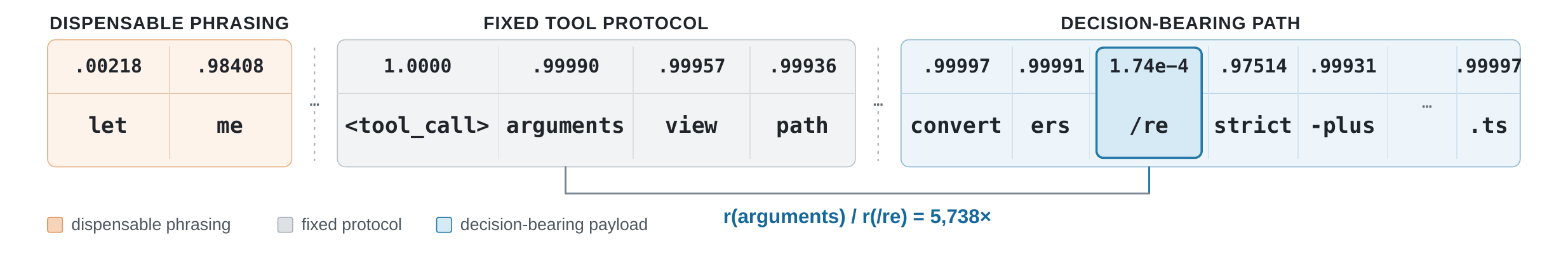}
    \caption{A typical case of DFT's supervision allocation. The value above each token is its target-token probability $p_t=p_{\theta,t}(y_t)$. DFT weights each token's cross-entropy by the stop-gradient of this value, $w_t=\operatorname{sg}[p_t]$; the per-token gradient-norm ratio then equals this weight, $\lVert\nabla_\theta \ell_t^{\mathrm{DFT}}\rVert_2 / \lVert\nabla_\theta \ell_t^{\mathrm{SFT}}\rVert_2 = w_t$.}
    \label{fig:dft_case}
\end{figure}

\paragraph{Token probability is a misaligned signal for agent SFT.}
Our first attempt in this family is DFT~\citep{wu2025dft}: it assigns supervision strength token by token according to the model's probability, i.e., higher-probability tokens receive more supervision. However, evaluation of DFT on Qwen3.5 backbone only resolves \textbf{40} SWE-bench~\citep{jimenez2024swebench} questions, substantially lower than \textbf{54} resolutions for Base~\footnote{DFT uses the same training and evaluation setup in Section~\ref{sec:experiments}.}. Figure~\ref{fig:dft_case} illustrates two potential explanations: i) the tokens receiving the strongest supervision are usually dispensable words or fixed protocol segments, which carry little information about a turn's reasoning logic or action decision; ii) the low-probability tokens are a mixture of stylistic preferences and key agent behaviors, so down-weighting them by probability also suppresses the supervision that matters.

\begin{figure}[h]
    \centering
       \begin{subfigure}{0.4\textwidth}
           \centering
           \includegraphics[width=\textwidth]{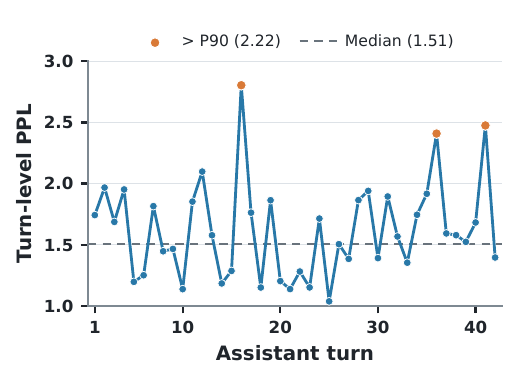}
           \caption{within a training trajectory.}
           \label{fig:turn_ppl_sequence}
       \end{subfigure}
       \begin{subfigure}{0.4\textwidth}
           \centering
           \includegraphics[width=\textwidth]{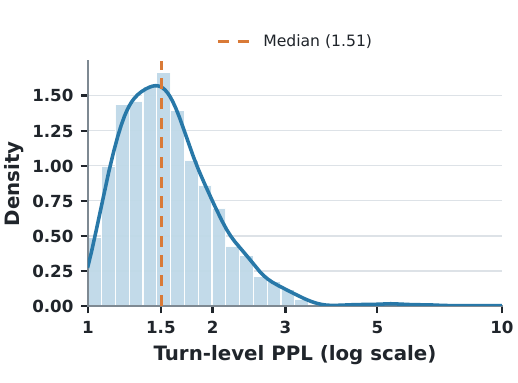}
           \caption{across the training data.}
           \label{fig:turn_ppl_distribution}
       \end{subfigure}
       \caption{Turn-level perplexity distribution of the base model on the training data.}
       \label{fig:turn_ppl}
\end{figure}

\paragraph{Agent turns carry distinguishable decision signals.}
In an agent trajectory, a turn couples reasoning with the action it produces, making an agent turn the smallest unit in which a complete decision appears. Moreover, a token-local score may miss how a decision is expressed, whereas a signal defined over the whole turn can better reflect the inherent characteristics of the demonstrated decision. Figure~\ref{fig:turn_ppl} exhibits this potential using average token perplexity (PPL) as the probing metric: within a single trajectory, per-turn perplexity spreads around a median of 1.51 up to a P90 of 2.22 (Figure~\ref{fig:turn_ppl_sequence}); across the training data, PPL is unimodal and right-skewed with a long tail (Figure~\ref{fig:turn_ppl_distribution}), showing distinguishable characteristics between turns.

However, a turn-level statistic by itself does not say which quantity should drive allocation. Raw perplexity, in particular, attributes a turn's surprise to everything the model does not expect, including stylistic expressions, and does not separate the part that follows from the turn's action decision. 
To this end, we derive a signal in the next section to isolate such components.

\subsection{Adaptive Supervision Objective: Privileged-Guided Supervised Fine-Tuning}
\label{sec:pgsft}

\begin{figure}[h]
    \centering
    \includegraphics[width=\textwidth]{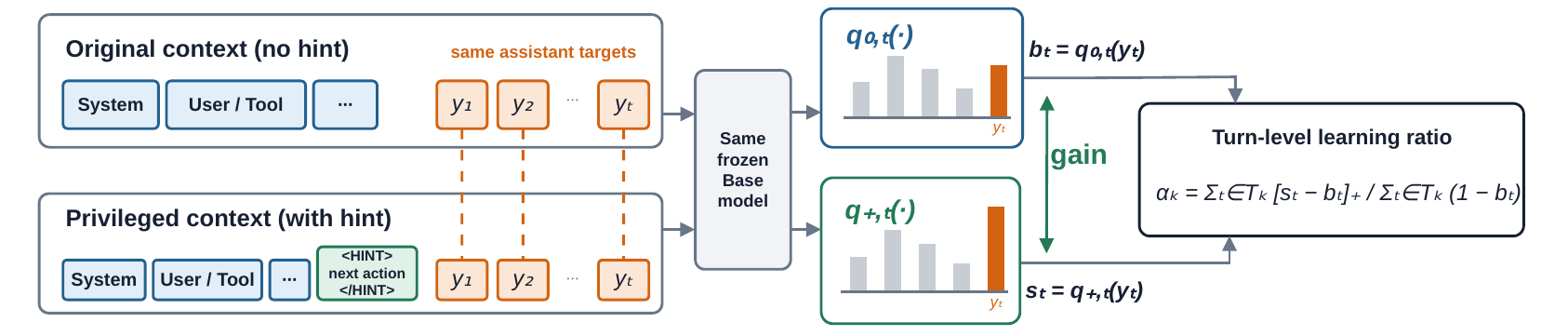}
    \caption{The same frozen base model scores aligned assistant tokens with and without an action hint. The resulting positive gain determines a shared turn-level coefficient $\alpha_k$.}
    \label{fig:privileged_conditioning}
\end{figure}

Motivated by preliminary insights, we leverage the idea of ``privileged context'' to construct a turn-level signal with this attribution. As illustrated in Figure~\ref{fig:privileged_conditioning}, a ``hint'' derived from the turn's demonstrated next action is inserted before its assistant tokens. PG-SFT then evaluates the \emph{same frozen base model} on the same target tokens under the ordinary prefix $x_t$ and the hint-augmented prefix $x_t^+$. Let $q_{+,t}(v)=p_0(v\mid x_t^+)$, $b_t=q_{0,t}(y_t)$, and $s_t=q_{+,t}(y_t)$.

Formally, for turn $k$ with supervised token positions $T_k$, we define the turn's positive gain $G_k$, its available probability room $H_k$, and its learning coefficient as
\begin{equation}
\begin{aligned}
G_k&=\sum_{t\in T_k}[s_t-b_t]_+, &\quad
H_k&=\sum_{t\in T_k}(1-b_t), &\quad
\alpha_k&=\begin{cases}G_k/H_k,&H_k>0,\\0,&H_k=0.\end{cases}
\end{aligned}
\label{eq:alpha}
\end{equation}
where $[z]_+=\max(z,0)$. Since $0\leq G_k\leq H_k$, the same $\alpha_k\in[0,1]$ applies to both reasoning and action tokens in the turn. Intuitively, $\alpha_k$ measures the fraction of the turn's probability room that revealing the action resolves. A turn whose content largely follows from its action scores high---the base model already supports such tokens once the decision is known, so fitting the turn mostly teaches the mapping from decision to realization. A turn that remains idiosyncratic even given the action scores low, and forcing its tokens would move the model away from the base distribution without acquiring decision-relevant behavior. Since the hint may reveal information unavailable at inference, such a coefficient is a measurement indicator of how action-determined a turn is, rather than the final supervision.

Based on this coefficient, we interpolate the target between the base distribution and the demonstrated token for every supervised token in the turn, as in Eq.~\ref{eq:fixed_target} but with a per-turn coefficient:
\begin{equation}
\boxed{
r_t^{\mathrm{PG}}(v)
=(1-\alpha_k)q_{0,t}(v)
+\alpha_k\mathbbm{1}[v=y_t],
\qquad t\in T_k.
}
\label{eq:pg_target}
\end{equation}
The target increases the demonstrated token's probability by $\alpha_k(1-b_t)$ while preserving the base model's relative probabilities among the other tokens; in particular, it does not distill the hint-conditioned distribution $q_{+,t}$. We train against this target with full-vocabulary forward KL~\citep{hinton2015distilling} with the frozen base model as teacher:
\begin{equation}
\mathcal{L}_{\mathrm{PG\text{-}SFT}}(\theta)
=\frac{1}{N_{\mathcal B}}
\sum_{i\in\mathcal B}\sum_{k\in i}\sum_{t\in T_k}
D_{\mathrm{KL}}\!\left(r_t^{\mathrm{PG}}\Vert p_{\theta,t}\right),
\qquad
N_{\mathcal B}=\sum_{i\in\mathcal B}\sum_{k\in i}|T_k|.
\label{eq:pg_loss}
\end{equation}
Here $\mathcal B$ is a batch of trajectories, $i$ indexes a trajectory, and $k$ indexes its assistant turns; gradients flow only through $p_\theta$.

Specifically, to separate \emph{where} supervision is allocated from its total target shift, we compare PG-SFT with a uniform coefficient
\begin{equation}
\bar\alpha=\frac{\sum_k G_k}{\sum_k H_k}
=\frac{\sum_k\alpha_k H_k}{\sum_k H_k},
\label{eq:matched_alpha}
\end{equation}
computed across the training turns. It matches the total increase in demonstrated-token probability induced by PG-SFT, $\sum_k\alpha_kH_k$. Consequently, this uniform target coincides with Base-KL at $\lambda=(1-\bar\alpha)/\bar\alpha$ for $0<\bar\alpha\leq1$.

%% file: sections/experiments.tex
\section{Experiments}
\label{sec:experiments}


\subsection{Experimental Setup}
\label{sec:exp_setup}
We evaluate the acquisition--retention trade-off among the supervision objectives; evaluation metrics, benchmark construction, and training hyperparameters are in Appendices~\ref{app:eval_protocol}, \ref{app:swebench_set},
and~\ref{app:training_details}.

\paragraph{Models and training data.}
We select two 4B-scale model families, Qwen3.5-4B~\cite{qwen2026qwen35} and Qwen3-4B-Thinking-2507~\cite{qwen2025qwen3}.
All fine-tuning objectives use the same offline agent trajectories sampled from the open-source Nemotron-SFT-SWE-v3 dataset~\cite{nvidia2026nemotronswev3}; for Qwen3.5-4B, the main experiments use 1K randomly sampled OpenHands~\cite{wang2024openhands} trajectories; for Qwen3-4B-Thinking-2507, we increase the training set to 5K to observe differentiable behavior after SFT. 

\paragraph{Benchmarks.} SWE-bench Lite~\cite{jimenez2024swebench} share similar trajectory format as the training data and serves as the primary target-task benchmark. GPQA Diamond~\cite{rein2024gpqa}, BFCL v4~\cite{patil2025bfcl}, and LiveCodeBench~\cite{jain2025livecodebench} evaluate capabilities retained during fine-tuning, from the aspects of general reasoning, multi-turn tool use, and code generation. Given the limited capability of 4B-scale backbones, some benchmarks are evaluated on their easier subsets to keep the scores informative. We report the \textbf{pass@3} metric~\cite{chen2021codex} to alleviate the impact of random sampling of agentic tasks, which also aligns with the mode coverage effect of SFT~\cite{gu2024minillm}.

\paragraph{Baselines.}
We evaluate five models for each backbone:
\textbf{Base}, \textbf{SFT}, \textbf{SFT+Base-KL}, \textbf{PSFT}, and \textbf{PG-SFT}, our adaptive objective described in Section~\ref{sec:pgsft}.
The four fine-tuning objectives observe exactly the same training data and the same supervised assistant tokens.
PSFT uses its published trust-region radius and optimization budget (Appendix~\ref{app:training_details}).

\paragraph{Hardware.} All training and evaluation used a single server with one Intel Xeon Platinum 8480C CPU, two NVIDIA RTX PRO 6000 Blackwell GPUs, and 512~GB of host memory.

\subsection{Target Adaptation and Capability Retention}
\label{sec:main_results}

Table~\ref{tab:main_results} reports the absolute success counts and two aggregate statistics: \textbf{Macro Avg}, the equal-weight average of the four benchmark success rates, and \textbf{$\Delta$ vs.\ Base}, its change over the Base model in accuracy percentage points (pp).

\begin{table*}[h]
\centering
\small
\setlength{\tabcolsep}{4pt}
\begin{tabular}{llcccccc}
\toprule
& &
\multicolumn{1}{c}{\textbf{Target Adaptation}} &
\multicolumn{3}{c}{\textbf{Cross-Benchmark Capability}} &
\multicolumn{2}{c}{\textbf{Overall}} \\
\cmidrule(lr){3-3}\cmidrule(lr){4-6}\cmidrule(lr){7-8}
\textbf{Backbone} &
\textbf{Objective} &
\textbf{SWE-bench} $\uparrow$ &
\textbf{GPQA} $\uparrow$ &
\textbf{BFCL} $\uparrow$ &
\textbf{LCB} $\uparrow$ &
\textbf{\shortstack{Macro\\Avg}} $\uparrow$ &
\textbf{\shortstack{$\Delta$ vs.\\Base}} $\uparrow$ \\
& & /89 & /198 & /200 & /120 & (\%) & (pp) \\
\midrule

\rowcolor{baserow}
\multirow{5}{*}{Qwen3.5-4B}
& Base              & 54 & 151 & 111 & 60 & 60.61 & 0.00 \\
& SFT               & \textbf{59} & 133 & 100 & 48 & 55.87 & \(-4.74\) \\
& SFT+Base-KL       & \underline{57} & 130 & 105 & 44 & 54.72 & \(-5.89\) \\
& PSFT              & 53 & \underline{142} & \textbf{112} & \textbf{63} & 59.94 & \(-0.67\) \\
\rowcolor{oursrow}
& \textbf{PG-SFT}            & \textbf{59} & \textbf{150} & \underline{111} & \underline{61} & \textbf{62.10} & \textbf{\(+1.49\)} \\

\midrule

\rowcolor{baserow}
\multirow{5}{*}{\shortstack{Qwen3-4B-\\Thinking}}
& Base              & 18 & 131 & 102 & 92 & 53.51 & 0.00 \\
& SFT               & \textbf{31} & 127 & 48 & 82 & 47.83 & \(-5.69\) \\
& SFT+Base-KL       & 16 & \textbf{133} & 83 & \underline{87} & 48.10 & \(-5.41\) \\
& PSFT              & 12 & 125 & \underline{108} & \textbf{88} & 50.99 & \(-2.53\) \\
\rowcolor{oursrow}
& \textbf{PG-SFT}            & \underline{27} & \underline{131} & \textbf{110} & 85 & \textbf{55.58} & \textbf{\(+2.07\)} \\

\bottomrule
\end{tabular}
\caption{
Target adaptation and cross-benchmark capability retention. Macro Avg averages the four per-benchmark success rates with equal weight; $\Delta$ vs.\ Base is its change over the Base model in accuracy percentage points (pp), computed within each backbone.
}
\label{tab:main_results}
\end{table*}

Results reveal a clear tension between fitting the target workflow and retaining capabilities already present in the Base model.
Standard SFT improves the target task but regresses on all three non-target benchmarks.
Adding a Base-KL regularizer reduces distributional deviation during training, but does not alleviate these downstream regressions.
Thus, globally encouraging the model to remain close to its initial distribution is not sufficient to preserve capabilities.

PSFT achieves one of the worst target-task results among the SFT objectives while retaining the most non-target capability among the baselines, aligning with findings that a trust-region constraint improves retention~\citep{zhu2026psft}; it nevertheless fails to learn effectively from the demonstrations.

PG-SFT exhibits a different acquisition--retention profile.
It resolves \(59\) of the 89 instances (\(66.3\%\)) on Qwen3.5, the highest target-task result we measure, while staying within a few instances of Base on non-target benchmarks. On Qwen3, PG-SFT gives up some target-task performance in exchange for stronger retention on non-target tasks.
This balanced profile also holds in aggregate: PG-SFT attains the highest Macro Avg on both backbones (\(62.10\%\) and \(55.58\%\)) and is the only fine-tuning objective that improves over its own Base model (\(+1.49\) pp and \(+2.07\) pp), whereas every other objective loses capability overall (\(-0.67\) to \(-5.89\) pp).

Despite being trained on the same offline demonstrations, objectives can induce substantially different changes outside the target task: stronger imitation does not necessarily translate into a better overall adaptation outcome, and a fixed Base-KL constraint does not reliably determine which existing behaviors should be preserved. To understand factors that may contribute to these differences, we next examine how each objective influences the learning process and distribution updates.



\subsection{How Much Do Objectives Learn from the Data?}
\label{sec:offline_analysis}

We evaluate the distribution alignment of each fine-tuned model with the same 120 held-out agent trajectories from the Nemotron-SFT-SWE-v3 dataset.
For all supervised assistant tokens, we measure their negative log-likelihood (NLL) and the forward KL divergence \(D_{\mathrm{KL}}(p_0\Vert p_\theta)\) between the frozen Base distribution and the fine-tuned model, i.e., the two statistics defined in Eq.~\ref{eq:metrics}. Results are summarized in Table~\ref{tab:offline_analysis}.

\begin{table*}[h]
\centering
\footnotesize
\setlength{\tabcolsep}{3pt}
\begin{tabular}{l
>{\columncolor{baserow}}c ccc >{\columncolor{oursrow}}c
>{\columncolor{baserow}}c ccc >{\columncolor{oursrow}}c}
\toprule
& \multicolumn{5}{c}{\textbf{Qwen3.5-4B}} &
\multicolumn{5}{c}{\textbf{Qwen3-4B-Thinking}} \\
\cmidrule(lr){2-6}\cmidrule(lr){7-11}
\textbf{Metric} &
\textbf{Base} & \textbf{SFT} & \textbf{\shortstack{SFT+\\Base-KL}} & \textbf{PSFT} & \textbf{PG-SFT} &
\textbf{Base} & \textbf{SFT} & \textbf{\shortstack{SFT+\\Base-KL}} & \textbf{PSFT} & \textbf{PG-SFT} \\
\midrule
\textbf{NLL} $\downarrow$
& 0.4302 & 0.3756 & 0.3767 & 0.4217 & 0.3968
& 0.7672 & 0.4903 & 0.5309 & 0.7423 & 0.5748 \\
\textbf{Base KL} $\downarrow$
& - & 0.0732 & 0.0431 & 0.0026 & 0.0080
& - & 0.2454 & 0.1105 & 0.0014 & 0.0342 \\
\bottomrule
\end{tabular}
\caption{
Teacher-forced analysis on held-out offline trajectories for both backbones. 
}
\label{tab:offline_analysis}
\end{table*}

\paragraph{KL regularization holds the drift, but cannot retain capability.} Standard SFT fits the demonstrations most aggressively and produces the largest observed distributional shift from the Base model.
SFT+Base-KL nearly halves this shift (\(0.0732\to0.0431\)) at essentially the same held-out NLL, yet the reduction does not translate into better capability retention (Table~\ref{tab:main_results}).

\paragraph{Learning conservatively withholds training towards the target.} PSFT fits the demonstrations least and stays closest to the Base distribution (Base KL \(0.0026\)), yet yields the worst target-task performance and the second-best capability profile (Table~\ref{tab:main_results}).

\paragraph{PG-SFT demonstrates clearer learning signals and distribution drift.} PG-SFT improves held-out NLL over the Base model while keeping Base KL at \(0.0080\), substantially below SFT and SFT+Base-KL: it absorbs information from the data, but more conservatively than SFT.

Interpreting these results together, if learning conservatively is sufficient to preserve capabilities, would uniformly constraining the learning intensity at the fixed level used by PG-SFT yield the same acquisition--retention trade-off?


\subsection{Is Conservative Learning Sufficient?}
\label{sec:matched_intensity}

We test this hypothesis with a \textbf{matched-intensity control}.
PG-SFT assigns a turn-dependent learning intensity \(\alpha_k\) while keeping the target on the Base-to-demonstration path of Eq.~\ref{eq:pg_target}; weighting each turn by its supervised token count \(|T_k|\), the average intensity over our Qwen3.5 training set is \(\bar{\alpha}_{\mathrm{token}}\approx 0.35\).
We therefore construct a fixed objective \(\mathcal{L}_{\mathrm{Fixed}} = 0.35\,\mathcal{L}_{\mathrm{CE}} + 0.65\,D_{\mathrm{KL}}(q_0\Vert p_\theta)\), i.e., the same intensity \(\alpha=0.35\) for every supervised token.
The control applies one learning intensity everywhere, whereas PG-SFT determines \(\alpha_k\) separately for each assistant turn from the privileged signal.

\begin{wraptable}{r}{0.4\textwidth}
\centering
\small
\vspace{0pt}
\begin{tabular}{lcc}
\toprule
\textbf{Objective} &
\textbf{NLL}  &
\textbf{Base KL}  \\
\midrule
Fixed intensity & 0.4006 & 0.0057 \\
PG-SFT          & 0.3968 & 0.0080 \\
\bottomrule
\end{tabular}
\caption{
Offline statistics for PG-SFT v.s. the matched fixed-intensity control.
}
\label{tab:matched_offline}
\end{wraptable}

As shown in Table~\ref{tab:matched_offline}, the fixed-intensity model closely matches PG-SFT's aggregate offline statistics (NLL \(0.4006\) vs.\ \(0.3968\); Base KL \(0.0057\) vs.\ \(0.0080\)): the control learns approximately the same teacher-forced drift while remaining at least as close to the Base distribution.

However, similar aggregate statistics do not produce the same downstream behavior.
As shown in Table~\ref{tab:matched_downstream}, the fixed-intensity model drops from \(59/89\) to \(50/89\) relative to PG-SFT on the SWE-bench subset and from \(150/198\) to \(140/198\) on GPQA, while BFCL stays at a similar level and LiveCodeBench even moves in the opposite direction.
These divergent outcomes confirm the significance of turn-adaptive supervision. PG-SFT's turn-specific intensity concentrates updates on turns whose content is resolved by the action hint and spares general-purpose tokens, producing selective changes that a single global coefficient cannot reproduce. 

\begin{table}[h]
\centering
\small
\setlength{\tabcolsep}{5pt}
\begin{tabular}{lcccc}
\toprule
\textbf{Objective} &
\textbf{SWE-bench} /89 &
\textbf{GPQA} /198 &
\textbf{BFCL} /200 &
\textbf{LCB} /120 \\
\midrule
Base              & 54 & 151 & 111 & 60 \\
Fixed intensity & 50 & 140 & 114 & 72 \\
PG-SFT          & 59 & 150 & 111 & 61 \\
\bottomrule
\end{tabular}
\caption{ Downstream performance of PG-SFT vs.\ the matched fixed-intensity control.}
\label{tab:matched_downstream}
\end{table}

\begin{wrapfigure}{r}{0.45\textwidth}
\centering
\includegraphics[width=\linewidth]{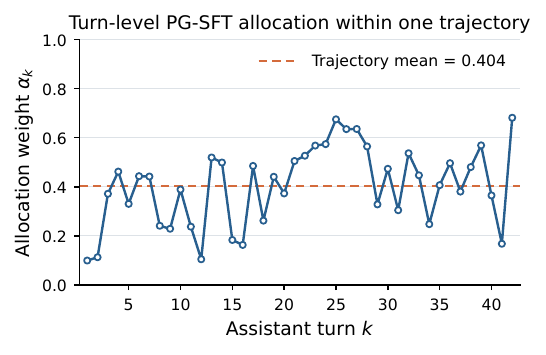}
\caption{PG-SFT learning intensity $\alpha$.}
\label{fig:turn-level-pg-sft-alpha}
\end{wrapfigure}

\paragraph{Why turn-adaptive supervision matters?} The observed PG-SFT coefficients are highly non-uniform across turns (Figure~\ref{fig:turn-level-pg-sft-alpha}): the turn-level median is \(0.244\), the 90th percentile \(0.607\), and the token-weighted mean near \(0.35\).
Combined with Tables~\ref{tab:matched_offline} and~\ref{tab:matched_downstream}, this indicates the significance of \emph{which supervised decisions are allowed to move away from the Base distribution}.

Overall, the experiments support four observations: i) stronger fitting of the training data may accompany substantial regressions in untrained capabilities; ii) a fixed Base-KL penalty reduces distributional drift but does not guarantee capability retention; iii) restricting update magnitudes, as PSFT does, keeps the model closest to the Base distribution yet commits to neither target acquisition nor non-target retention; iv) matching PG-SFT's average learning intensity does not reproduce its downstream behavior, motivating adaptive allocation of learning pressure across supervised decisions rather than a single global rule.

%% file: sections/conclusion.tex
\section{Discussion and Conclusion}
\label{sec:conclusion}

\paragraph{Correlations between turn coefficient and turn length.}
$\alpha_k$ is an average value, and thus could strongly correlate with the number of tokens in a turn. We analyzed 5{,}503 assistant turns from 120 held-out trajectories to understand their relationship. While the turn length correlates with $G_k$ and $H_k$ individually ($r=0.66$ and $0.75$), the two dependencies largely cancel in the ratio, leaving only a weak correlation with $\alpha_k$ (Pearson $r=0.225$, cluster-bootstrap 95\% CI $[0.184,\,0.269]$, univariate $R^2\approx5\%$). Therefore, despite the turn-level averaging of $\alpha_k$, the coefficient reflects more on what a turn contains rather than how many tokens it has.

\paragraph{Scope \& Limitations.}
Our distributional measurements are empirical, and cannot establish a causal link between drift and retention. Our evaluation covers two related 4B-scale backbones, a single agent domain, and a limited number of training runs. Therefore, we cannot promise our findings can be extended to training data and model size at any scale: although fine-tuning regressions have been documented in larger models~\citep{qi2023safety,luo2023forgetting}, data-selection effects are capacity-dependent~\citep{dang2026capacity}, so neither a pure capacity account nor a pure objective account can be excluded. Finally, privileged scoring and full-vocabulary KL add computation over ordinary SFT, although the privileged scores are computed once for a dataset and a model.

\paragraph{Conclusion.}
This work studies how supervision-objective design affects the acquisition--retention trade-off in offline agent fine-tuning. Comparing standard SFT, Base-KL, PSFT, and PG-SFT on identical trajectories shows that target fitting, base proximity, and downstream retention are distinct: SFT forgets non-target capabilities; Base-KL reduces drift without reliable retention gains; PSFT stays near the base model but learns little. PG-SFT improves the trade-off through adaptive, turn-level supervision. Balancing acquisition and retention thus depends not merely on anchoring to the base model, but on deciding where and how strongly to depart from it.

%% file: sections/appendix.tex

\section{Implementation details}\label{app:implementation}

\paragraph{Distributed normalization.} The objective in
Eq.~\ref{eq:pg_loss} is defined once per optimizer update but computed per
microbatch. With $R$ data-parallel ranks and $K_k$ microbatches actually executed
per rank in update $k$, the implementation injects the global divisor
\begin{equation}\label{eq:divisor}
d_k=\frac{N_k}{RK_k},
\qquad
\ell_{r,b}=\frac{1}{d_k}\sum_{(i,t)\in\mathcal B_{r,b}}
D_{\mathrm{KL}}\!\left(r_t\Vert p_{\theta,t}\right),
\end{equation}
so that averaging over ranks and microbatches recovers the objective exactly,
\begin{equation}\label{eq:equiv}
\frac{1}{R}\sum_{r=1}^{R}\frac{1}{K_k}\sum_{b=1}^{K_k}
\nabla_\theta\,\ell_{r,b}
=\nabla_\theta\,\mathcal L_{\mathrm{PG\text{-}SFT}}.
\end{equation}
Here $N_k$ is the number of supervised assistant tokens in update $k$
(cf.~$N_{\mathcal B}$ in Eq.~\ref{eq:pg_loss} for a full batch), and
$\mathcal B_{r,b}$ is the set of token positions in microbatch $b$ on rank $r$.
Trailing updates with fewer than $K_k$ microbatches divide by the number actually
executed.

\paragraph{Exact KL without materializing the batch.} Both logarithms in
Eq.~\ref{eq:pg_loss} require the full vocabulary at every supervised position, which
a standard forward pass would materialize as a tensor of size $N_k\times|V|$. The
implementation instead receives the student hidden states $H$ together with the
targets $r_t$ built from the frozen base lane, and projects the hidden states in
token-row chunks $C$,
\begin{equation}\label{eq:chunk}
Z_C=H_CW_{\mathrm{out}}^\top,
\qquad
p_C=\mathrm{softmax}(Z_C),
\end{equation}
accumulating the exact KL chunk by chunk, so the logit workspace scales with
$O(|C||V|)$ rather than $O(N_k|V|)$ while the value and gradient are those of the
unchunked objective. The custom backward for the student path is
\begin{equation}\label{eq:chunk-bwd}
\frac{\partial \ell}{\partial H_C}
=\frac{1}{d_k}\left(p_C-r_C\right)W_{\mathrm{out}},
\end{equation}
the gradient of Eq.~\ref{eq:pg_loss} composed with the output projection.

\paragraph{Cost.} Relative to SFT, a PG-SFT update additionally runs the frozen
base model over the ordinary and privileged contexts, constructs a
full-vocabulary target at every supervised position, and retains the intermediate
states that the chunked backward needs. The clipped-ratio objective instead
requires the demonstrated-token probabilities of the reference policy that
defines its ratio, in addition to the student forward pass. We therefore do not
claim equal compute for the compared objectives, which are matched on data and
supervised tokens only (Section~\ref{sec:experiments}). Diagnostics such as
negative log-likelihood, perplexity, and entropy on the demonstrated tokens are
logged for analysis and are not part of any objective.





\section{Training contract}\label{app:training_details}

\paragraph{Training data.} The offline trajectories are sampled from the
open-source Nemotron-SFT-SWE-v3 dataset~\cite{nvidia2026nemotronswev3} and follow
the OpenHands~\cite{wang2024openhands} agent format, which matches the trajectory format of SWE-bench. The
Qwen3.5-4B experiments use 1{,}000 randomly sampled trajectories; the
Qwen3-4B-Thinking-2507 experiments use 4{,}992 trajectories, because
differences between objectives were not measurable at the smaller size for that
backbone.

Table~\ref{tab:training_contract} lists the settings shared by all compared
objectives on the Qwen3.5-4B backbone and what each objective adds; only the
supervision objective and its own hyperparameters differ.
Table~\ref{tab:training_contract_qwen3} reports the corresponding contract for
the Qwen3-4B-Thinking-2507 backbone.

\begin{table}[ht]
\centering
\small
\setlength{\tabcolsep}{5pt}
\begin{tabular}{ll}
\toprule
\textbf{Shared setting} & \textbf{Value} \\
\midrule
Backbone & Qwen3.5-4B \\
Training trajectories & 1{,}000 offline agent trajectories \\
Supervised region & complete assistant generation mask \\
Holdout panel & 120 fixed trajectories \\
Seed & 42 \\
Precision & BF16 \\
Adaptation & LoRA~\citep{hu2021lora} rank 32, $\alpha=64$, dropout $0$ on all linear layers except \texttt{lm\_head} \\
Optimization & AdamW, learning rate $5\times10^{-5}$, 5 warmup steps, gradient clip $1.0$ \\
Batch and schedule & global batch $32$; cosine decay over a 64-step horizon, $32$ steps executed \\
Checkpoint & fixed final-step checkpoint (LoRA-merged weights) \\
\midrule
\textbf{Objective} & \textbf{Objective-specific settings} \\
\midrule
SFT & cross-entropy on the demonstrated tokens (Eq.~\ref{eq:sft}) \\
SFT+Base-KL & SFT plus forward $\mathrm{KL}(p_0\Vert p_\theta)$ with $\lambda=0.1$ (Eq.~\ref{eq:base_kl}) \\
PG-SFT & frozen base lanes, temperature $1$, turn-level $\alpha_k$ of Eq.~\ref{eq:alpha} \\
PSFT & clipped surrogate of Eq.~\ref{eq:psft} with $\epsilon=0.28$; 3 epochs \\
     & (96 optimizer steps) at learning rate $10^{-6}$ \\
Fixed intensity & $0.35\,\mathcal{L}_{\mathrm{CE}}+0.65\,D_{\mathrm{KL}}(q_0\Vert p_\theta)$ \\
                & (Section~\ref{sec:matched_intensity}); step-32 merged checkpoint \\
\bottomrule
\end{tabular}
\caption{
Training contract of the compared objectives on the Qwen3.5-4B backbone.
The SFT, SFT+Base-KL, PG-SFT, and fixed-intensity arms share their adaptation
and optimization configuration; the PSFT arm follows its own schedule and is
therefore reported as a reference point rather than as a budget-matched
comparison (Section~\ref{sec:exp_setup}).
}
\label{tab:training_contract}
\end{table}

The $\lambda=0.1$ coefficient of the SFT+Base-KL arm places its fixed target
(Eq.~\ref{eq:fixed_target}) at
$\alpha=1/(1+\lambda)\approx0.91$, a
much weaker reduction of supervision than the matched fixed-intensity control
($\alpha=0.35$) of Section~\ref{sec:matched_intensity}.

\begin{table}[ht]
\centering
\small
\setlength{\tabcolsep}{5pt}
\begin{tabular}{ll}
\toprule
\textbf{Shared setting} & \textbf{Value} \\
\midrule
Backbone & Qwen3-4B-Thinking-2507 \\
Training trajectories & 4{,}992 offline agent trajectories \\
Supervised region & complete assistant generation mask \\
Holdout panel & 120 fixed trajectories \\
Seed & 42 \\
Precision & BF16 \\
Adaptation & LoRA rank 32, $\alpha=64$, dropout $0$ on all linear layers except \texttt{lm\_head} \\
Optimization & AdamW, learning rate $5\times10^{-5}$, 5 warmup steps, gradient clip $1.0$ \\
Batch & global batch $32$ \\
Checkpoint & fixed final-step checkpoint (LoRA-merged weights) \\
\midrule
\textbf{Objective} & \textbf{Objective-specific settings} \\
\midrule
SFT & cross-entropy on the demonstrated tokens (Eq.~\ref{eq:sft}); \\
    & 156 steps (one epoch), cosine decay \\
SFT+Base-KL & SFT plus forward $\mathrm{KL}(p_0\Vert p_\theta)$ with $\lambda=0.1$ (Eq.~\ref{eq:base_kl}); \\
    & 32 steps, cosine decay \\
PG-SFT & frozen base lanes, temperature $1$, turn-level $\alpha_k$ of Eq.~\ref{eq:alpha}; \\
    & 156 steps (one epoch), cosine decay \\
PSFT & clipped surrogate of Eq.~\ref{eq:psft} with $\epsilon=0.28$; 3 epochs \\
     & (96 optimizer steps) at learning rate $10^{-6}$ \\
\bottomrule
\end{tabular}
\caption{
Training contract of the compared objectives on the Qwen3-4B-Thinking-2507
backbone. Adaptation, optimization, seed, and data processing match
Table~\ref{tab:training_contract}; only the training-set size and the
per-objective optimization budgets differ. In particular, the SFT+Base-KL arm
executes 32 steps rather than a full epoch, so it is not budget-matched to the
SFT and PG-SFT arms on this backbone.
}
\label{tab:training_contract_qwen3}
\end{table}

In Table~\ref{tab:training_contract_qwen3}, the SFT+Base-KL arm uses the same
coefficient $\lambda=0.1$ as on Qwen3.5-4B but trains for 32 steps instead of
one epoch (156 steps); on this backbone its comparison against the SFT and
PG-SFT arms is therefore not budget-matched, in the same sense as the PSFT
schedule.

\section{Evaluation protocol}\label{app:eval_protocol}

\paragraph{Benchmark sets.} GPQA Diamond~\cite{rein2024gpqa} uses all 198 questions in the frozen set. BFCL v4~\cite{patil2025bfcl} uses a fixed 200-instance subset of the multi-turn categories (50 per category) scored by the official \texttt{bfcl-eval} harness; this subset is a development split of the benchmark rather than its official test split. LiveCodeBench~\cite{jain2025livecodebench} uses a frozen 120-instance stratified subset of release v6. SWE-bench~\cite{jimenez2024swebench} uses the 89-instance set of Appendix~\ref{app:swebench_set} with the official SWE-bench harness verification.

\paragraph{Metric.} The target-task column reports \textbf{pass@3}: an instance is counted as resolved when the harness resolves at least one of its attempts. The attempt schedule of each subset of the 89 instances is given in Appendix~\ref{app:swebench_set}.

\paragraph{Generation budget.} SWE-bench rollouts use temperature 1, at most 8192 new tokens per round, at most 150 rounds, and a context of 131072 tokens.  GPQA is generated at temperature 0.6 with top-$p$ 0.95, top-$k$ 20, and at most 32768 new tokens in the first pass; instances whose first output is truncated at that budget are re-run with a 65536-token budget. BFCL is generated at temperature 0 with seed 42, at most 8192 new tokens per request, and a context of 262144 tokens.

\paragraph{Scoring.} GPQA is graded by extracting the final answer under a fixed answer rule: an example counts as correct when the correct option is identifiable from the response, and examples without an identifiable final answer count as incorrect. SWE-bench, BFCL, and LiveCodeBench use their task-specific execution-based evaluators.

\section{SWE-bench evaluation set}\label{app:swebench_set}

The target-task results in Table~\ref{tab:main_results} are evaluated on a frozen set of 89 SWE-bench Lite instances. We construct this set to improve discriminability: on a uniformly random sample of Lite, the two backbones often achieve similar performance, providing limited signal for comparing training objectives. The final set combines a random subset with a difficulty-screened subset where the two backbones exhibit clearer separation.

\paragraph{Official Lite split.}
SWE-bench Lite is constructed from the SWE-bench test set through a filtering and sampling procedure that removes low-quality or ambiguous instances and selects 300 test instances.\footnote{\url{https://github.com/SWE-bench/SWE-bench/blob/v4.1.0/swebench/collect/make_lite/README.md}}
We follow the official harness and evaluation protocol; only the evaluated instance subset differs.

\paragraph{Random subset.}
The random subset is the frozen 50-instance sample used in our earlier experiments, drawn from the 300 Lite instances with \texttt{random.sample} (seed 42). It is not difficulty-screened or repository-stratified.

\paragraph{Difficulty-screened subset.}
The screened subset is constructed using external evaluation evidence without using any model evaluated in this paper. We aggregate public SWE-bench Lite evaluation records and repeated-sampling statistics to estimate instance-level solvability, remove known flaky instances, and select 50 instances with higher and more consistent resolution frequencies. The selected instances have average resolution frequencies of 62.1\% and 64.0\% under the two statistics, respectively. These values characterize the selected instances and should not be interpreted as model pass rates.

\paragraph{Union and scoring.}
The two subsets overlap on 11 instances, resulting in an 89-instance evaluation set after deduplication by \textit{instance\_id}. All instances use the same agent scaffold, prompt template, and output budget. For the random subset, we follow the previous evaluation protocol; for the screened subset, we additionally evaluate up to three attempts for unresolved instances. Since additional attempts are allocated conditionally based on previous outcomes, the reported pass@3 should be viewed as a comparison protocol rather than an unbiased estimate of a pre-registered three-sample evaluation. The resulting 89-instance scores are not directly comparable to SWE-bench Lite leaderboard numbers, which are reported on the full 300-instance test set.
\section{Future directions}\label{app:future}

Four directions follow directly from the limitations discussed in Section~\ref{sec:conclusion}. First, the design space of the hint itself is unexplored: its granularity, its source (the next action, environment feedback, or the final outcome), and counterfactual hints as a specificity check on the signal. Second, coefficients could be re-estimated against the current policy rather than the frozen base, tracking the diminishing returns of each turn as training proceeds. Third, turn-level allocation is orthogonal to rehearsal and to online reinforcement learning---where rollout outcomes provide privileged information essentially for free---and combining them is a natural next step. Fourth, comparing privileged gain against other turn-level signals (e.g., margin- or entropy-based scores) would isolate whether action attribution is necessary for the retention benefit.

%% file: references.bib
@article{zeng2023agenttuning,
  title={AgentTuning: Enabling Generalized Agent Capabilities for LLMs},
  author={Zeng, Aohan and others},
  journal={arXiv preprint arXiv:2310.12823},
  year={2023}
}

@inproceedings{li2017learning,
  title={Learning without Forgetting},
  author={Li, Zhizhong and Hoiem, Derek},
  booktitle={European Conference on Computer Vision (ECCV)},
  pages={614--629},
  year={2016}
}

@article{kirkpatrick2017overcoming,
  title={Overcoming catastrophic forgetting in neural networks},
  author={Kirkpatrick, James and others},
  journal={PNAS},
  year={2017}
}

@inproceedings{li2018l2sp,
  title={Explicit Inductive Bias for Transfer Learning with Convolutional Networks},
  author={Li, Xuhong and Grandvalet, Yves and Davoine, Franck},
  booktitle={ICML},
  year={2018}
}

@inproceedings{kemker2018measuring,
  title={Measuring catastrophic forgetting in neural networks},
  author={Kemker, Ronald and McClure, Marc and Abitino, Angelina and Hayes, Tyler and Kanan, Christopher},
  booktitle={Proceedings of the AAAI conference on artificial intelligence},
  volume={32},
  year={2018}
}

@inproceedings{ouyang2022instructgpt,
  author    = {Long Ouyang and Jeffrey Wu and Xu Jiang and Diogo Almeida and
               Carroll L. Wainwright and Pamela Mishkin and Chong Zhang and
               Sandhini Agarwal and Katarina Slama and Alex Ray and others},
  title     = {Training Language Models to Follow Instructions with Human Feedback},
  booktitle = {Advances in Neural Information Processing Systems (NeurIPS)},
  year      = {2022},
}

@misc{touvron2023llama2,
  author       = {Hugo Touvron and Louis Martin and Kevin Stone and Peter Albert and others},
  title        = {Llama 2: Open Foundation and Fine-Tuned Chat Models},
  year         = {2023},
  howpublished = {arXiv:2307.09288},
}

@inproceedings{yao2023react,
  author    = {Shunyu Yao and Jeffrey Zhao and Dian Yu and Nan Du and Izhak Shafran and
               Karthik Narasimhan and Yuan Cao},
  title     = {{ReAct}: Synergizing Reasoning and Acting in Language Models},
  booktitle = {International Conference on Learning Representations (ICLR)},
  year      = {2023},
}

@inproceedings{chen2024agentflan,
  author    = {Zehui Chen and Kuikun Liu and Qiuchen Wang and Wenwei Zhang and Jiangning Liu and
               Dahua Lin and Kai Chen and Feng Zhao},
  title     = {{Agent-FLAN}: Designing Data and Methods of Effective Agent Tuning for Large Language Models},
  booktitle = {Findings of the Association for Computational Linguistics: ACL 2024},
  year      = {2024},
}

@misc{xu2026masktarget,
  author       = {Runze Xu and Arpit Garg and Hemanth Saratchandran and Simon Lucey},
  title        = {Mask the Target: A Plug-and-Play Regularizer Against {LoRA} Forgetting},
  year         = {2026},
  howpublished = {arXiv:2605.29498},
}

@misc{wang2026memsft,
  author       = {Jiarui Wang and Xiang Shi and Jiaqi Cao and Rubin Wei and others},
  title        = {{MemSFT}: Mitigating Alignment Tax with an External Parametric Memory},
  year         = {2026},
  howpublished = {arXiv:2607.25614},
}

@misc{luo2023forgetting,
  author       = {Yun Luo and Zhen Yang and Fandong Meng and Yafu Li and Jie Zhou and Yue Zhang},
  title        = {An Empirical Study of Catastrophic Forgetting in Large Language Models During Continual Fine-tuning},
  year         = {2023},
  howpublished = {arXiv:2308.08747},
}

@misc{lin2025sft,
  author       = {Jiacheng Lin and Zhongruo Wang and Kun Qian and Tian Wang and Arvind Srinivasan and Hansi Zeng and others},
  title        = {{SFT} Doesn't Always Hurt General Capabilities: Revisiting Domain-Specific Fine-Tuning in {LLMs}},
  year         = {2025},
  howpublished = {arXiv:2509.20758},
}

@misc{biderman2024lora,
  author       = {Dan Biderman and Jose Gonzalez Ortiz and Jacob Portes and Mansheej Paul and others},
  title        = {{LoRA} Learns Less and Forgets Less},
  year         = {2024},
  howpublished = {arXiv:2405.09673},
}

@misc{huan2025math,
  author       = {Maggie Huan and Yuetai Li and Tuney Zheng and Xiaoyu Xu and Seungone Kim and Minxin Du and Radha Poovendran and Graham Neubig and Xiang Yue},
  title        = {Does Math Reasoning Improve General {LLM} Capabilities? Understanding Transferability of {LLM} Reasoning},
  year         = {2025},
  howpublished = {arXiv:2507.00432},
}

@inproceedings{wang2024promot,
  author    = {Yihan Wang and Si Si and Daliang Li and Michal Lukasik and Felix Yu and Cho-Jui Hsieh and Inderjit S. Dhillon and Sanjiv Kumar},
  title     = {Two-stage {LLM} Fine-tuning with Less Specialization and More Generalization},
  booktitle = {International Conference on Learning Representations (ICLR)},
  year      = {2024},
}

@inproceedings{qi2023safety,
  author    = {Xiangyu Qi and Yi Zeng and Tinghao Xie and Pin-Yu Chen and
               Ruoxi Jia and Prateek Mittal and Peter Henderson},
  title     = {Fine-tuning Aligned Language Models Compromises Safety, Even When Users Do Not Intend To!},
  booktitle = {International Conference on Learning Representations (ICLR)},
  year      = {2024},
}

@inproceedings{zhang2025grape,
  author    = {Dylan Zhang and Qirun Dai and Hao Peng},
  title     = {The Best Instruction-Tuning Data are Those That Fit},
  booktitle = {Advances in Neural Information Processing Systems (NeurIPS)},
  year      = {2025},
}

@inproceedings{yang2024sdft,
  author    = {Zhaorui Yang and others},
  title     = {Self-Distillation Bridges Distribution Gap in Language Model Fine-Tuning},
  booktitle = {Proceedings of the 62nd Annual Meeting of the Association for Computational Linguistics (ACL)},
  year      = {2024},
}

@inproceedings{li2025small,
  author    = {Yuetai Li and others},
  title     = {Small Models Struggle to Learn from Strong Reasoners},
  booktitle = {Findings of the Association for Computational Linguistics: ACL 2025},
  year      = {2025},
}

@misc{dang2026capacity,
  author       = {Cuong Dang and Hoang Anh Just and Ruoxi Jia},
  title        = {Capacity-Dependent Effects of Data Selection for Reasoning},
  year         = {2026},
  howpublished = {arXiv:2608.13721},
}

@inproceedings{li2024ifd,
  author    = {Ming Li and others},
  title     = {From Quantity to Quality: Boosting {LLM} Performance with Self-Guided Data Selection for Instruction Tuning},
  booktitle = {Proceedings of the 2024 Conference of the North American Chapter of the Association for Computational Linguistics (NAACL)},
  year      = {2024},
}

@inproceedings{liu2024deita,
  author    = {Wei Liu and others},
  title     = {What Makes Good Data for Alignment? A Comprehensive Study of Automatic Data Selection in Instruction Tuning},
  booktitle = {International Conference on Learning Representations (ICLR)},
  year      = {2024},
}

@misc{liu2024selectit,
  author       = {Liangxin Liu and others},
  title        = {{SelectIT}: Selective Instruction Tuning for {LLM}s via Uncertainty-Aware Self-Reflection},
  year         = {2024},
  howpublished = {arXiv:2402.16705},
}

@misc{chen2023alpagasus,
  author       = {Lichang Chen and others},
  title        = {{AlpaGasus}: Training a Better Alpaca with Fewer Data},
  year         = {2023},
  howpublished = {arXiv:2307.08701},
}

@inproceedings{xia2025rethinking,
  author    = {Tingyu Xia and others},
  title     = {Rethinking Data Selection at Scale: Random Selection is Almost All You Need},
  booktitle = {Findings of the Association for Computational Linguistics: EMNLP 2025},
  year      = {2025},
}

@inproceedings{li2024reflection,
  author    = {Ming Li and others},
  title     = {Selective Reflection-Tuning: Student-Selected Data Recycling for {LLM} Instruction-Tuning},
  booktitle = {Findings of the Association for Computational Linguistics: ACL 2024},
  year      = {2024},
}

@inproceedings{wu2025dft,
  author    = {Yongliang Wu and others},
  title     = {On the Generalization of {SFT}: A Reinforcement Learning Perspective with Reward Rectification},
  booktitle = {International Conference on Learning Representations (ICLR)},
  year      = {2026},
}

@misc{zhang2026idft,
  author       = {Miaosen Zhang and others},
  title        = {Towards On-Policy {SFT}: Distribution Discriminant Theory and its Applications in {LLM} Training},
  year         = {2026},
  howpublished = {arXiv:2602.12222},
}

@inproceedings{zhu2025asft,
  author    = {He Zhu and others},
  title     = {Anchored Supervised Fine-Tuning},
  booktitle = {International Conference on Learning Representations (ICLR)},
  year      = {2026},
}

@inproceedings{zhu2026psft,
  author    = {Wenhong Zhu and Ruobing Xie and Rui Wang and Xingwu Sun and Di Wang and Pengfei Liu},
  title     = {Proximal Supervised Fine-Tuning},
  booktitle = {International Conference on Learning Representations (ICLR)},
  year      = {2026},
}

@misc{zhou2026cadft,
  author       = {Yucheng Zhou and others},
  title        = {Compatibility-Aware Dynamic Fine-Tuning for Large Language Models},
  year         = {2026},
  howpublished = {arXiv:2606.11206},
}

@misc{sabbaghi2026infosft,
  author       = {Mahdi Sabbaghi and George J. Pappas and Adel Javanmard and Hamed Hassani},
  title        = {{InfoSFT}: Learn More and Forget Less with Information-Aware Token Weighting},
  year         = {2026},
  howpublished = {arXiv:2605.14967},
}

@misc{hinton2015distilling,
  author       = {Geoffrey Hinton and Oriol Vinyals and Jeff Dean},
  title        = {Distilling the Knowledge in a Neural Network},
  year         = {2015},
  howpublished = {arXiv:1503.02531},
}

@inproceedings{lopezdpaz2016unifying,
  author    = {David Lopez-Paz and L\'eon Bottou and Bernhard Sch\"olkopf and Vladimir Vapnik},
  title     = {Unifying Distillation and Privileged Information},
  booktitle = {International Conference on Learning Representations (ICLR)},
  year      = {2016},
}

@article{vapnik2015lupi,
  author  = {Vladimir Vapnik and Rauf Izmailov},
  title   = {Learning Using Privileged Information: Similarity Control and Knowledge Transfer},
  journal = {Journal of Machine Learning Research},
  volume  = {16},
  number  = {61},
  pages   = {2023--2049},
  year    = {2015},
}

@inproceedings{yang2022privileged,
  author    = {Shuo Yang and Sujay Sanghavi and Holakou Rahmanian and Jan Bakus and S. V. N. Vishwanathan},
  title     = {Toward Understanding Privileged Features Distillation in Learning-to-Rank},
  booktitle = {Advances in Neural Information Processing Systems (NeurIPS)},
  pages     = {26658--26670},
  year      = {2022},
}

@inproceedings{aslam2023privileged,
  author    = {Muhammad Haseeb Aslam and Muhammad Osama Zeeshan and Marco Pedersoli and Alessandro L. Koerich and Simon Bacon and Eric Granger},
  title     = {Privileged Knowledge Distillation for Dimensional Emotion Recognition in the Wild},
  booktitle = {IEEE/CVF Conference on Computer Vision and Pattern Recognition Workshops (CVPRW)},
  pages     = {3338--3347},
  year      = {2023},
}

@misc{snell2022distillcontext,
  author       = {Charlie Snell and Dan Klein and Ruiqi Zhong},
  title        = {Learning by Distilling Context},
  year         = {2022},
  howpublished = {arXiv:2209.15189},
}

@inproceedings{agarwal2024gkd,
  author    = {Rishabh Agarwal and others},
  title     = {On-Policy Distillation of Language Models: Learning from Self-Generated Mistakes},
  booktitle = {International Conference on Learning Representations (ICLR)},
  year      = {2024},
}

@inproceedings{gu2024minillm,
  author    = {Yuxian Gu and Li Dong and Furu Wei and Minlie Huang},
  title     = {{MiniLLM}: Knowledge Distillation of Large Language Models},
  booktitle = {International Conference on Learning Representations (ICLR)},
  year      = {2024},
}

@misc{zhao2026opsd,
  author       = {Siyan Zhao and others},
  title        = {Self-Distilled Reasoner: On-Policy Self-Distillation for Large Language Models},
  year         = {2026},
  howpublished = {arXiv:2601.18734},
}

@misc{penaloza2026pidistill,
  author       = {Emiliano Penaloza and others},
  title        = {Privileged Information Distillation for Language Models},
  year         = {2026},
  howpublished = {arXiv:2602.04942},
}

@misc{shenfeld2025razor,
  author       = {Idan Shenfeld and Jyothish Pari and Pulkit Agrawal},
  title        = {{RL}'s Razor: Why Online Reinforcement Learning Forgets Less},
  year         = {2025},
  howpublished = {arXiv:2509.04259},
}

@misc{harne2026privilegedbiased,
  author       = {Sarthak Harne and others},
  title        = {Privileged, but Biased: How {PI}-Conditioned Teachers Break Self-Distillation},
  year         = {2026},
  howpublished = {arXiv:2608.04794},
}

@misc{kaur2026thinking,
  author       = {Simran Kaur and others},
  title        = {Rethinking On-Policy Self-Distillation for Thinking Models},
  year         = {2026},
  howpublished = {arXiv:2607.05184},
}

@misc{shrestha2026rethinkingpi,
  author       = {Samyak Shrestha and Alexander Tessier},
  title        = {Rethinking Privileged Information in On-Policy Self-Distillation},
  year         = {2026},
  howpublished = {arXiv:2608.18271},
}

@misc{zhao2026morepi,
  author       = {Xuyang Zhao and others},
  title        = {Is More Privileged Information Better? From Solution Traces to Problem-Solving Structure in Self-Distilled Reasoning},
  year         = {2026},
  howpublished = {arXiv:2608.01589},
}

@inproceedings{hu2021lora,
  author    = {Edward J. Hu and Yelong Shen and Phillip Wallis and Zeyuan Allen-Zhu and Yuanzhi Li and Shean Wang and Lu Wang and Weizhu Chen},
  title     = {{LoRA}: Low-Rank Adaptation of Large Language Models},
  booktitle = {International Conference on Learning Representations (ICLR)},
  year      = {2022},
}

@inproceedings{jimenez2024swebench,
  author    = {Carlos E. Jimenez and John Yang and Alexander Wettig and Shunyu Yao and
               Kexin Pei and Ofir Press and Karthik R. Narasimhan},
  title     = {{SWE-bench}: Can Language Models Resolve Real-World GitHub Issues?},
  booktitle = {International Conference on Learning Representations (ICLR)},
  year      = {2024},
}

@inproceedings{rein2024gpqa,
  author    = {David Rein and Betty Li Hou and Asa Cooper Stickland and Jackson Petty and
               Richard Yuanzhe Pang and Julien Dirani and Julian Michael and Samuel R. Bowman},
  title     = {{GPQA}: A Graduate-Level Google-Proof {Q\&A} Benchmark},
  booktitle = {First Conference on Language Modeling (COLM)},
  year      = {2024},
}

@inproceedings{patil2025bfcl,
  author    = {Shishir G. Patil and Huanzhi Mao and Fanjia Yan and Charlie Cheng-Jie Ji and
               Vishnu Suresh and Ion Stoica and Joseph E. Gonzalez},
  title     = {The Berkeley Function Calling Leaderboard ({BFCL}): From Tool Use to
               Agentic Evaluation of Large Language Models},
  booktitle = {International Conference on Machine Learning (ICML)},
  year      = {2025},
}

@inproceedings{jain2025livecodebench,
  author    = {Naman Jain and King Han and Alex Gu and Wen-Ding Li and Fanjia Yan and
               Tianjun Zhang and Sida Wang and Armando Solar-Lezama and Koushik Sen and
               Ion Stoica},
  title     = {{LiveCodeBench}: Holistic and Contamination Free Evaluation of Large
               Language Models for Code},
  booktitle = {International Conference on Learning Representations (ICLR)},
  year      = {2025},
}

@misc{nvidia2026nemotronswev3,
  author = {{NVIDIA}},
  title  = {{Nemotron-SFT-SWE-v3}: Software Engineering Instruction Tuning Data},
  year   = {2026},
  note   = {Agentic trajectories from the {OpenHands}, {SWE-agent}, and
            {mini-SWE-agent} harnesses for {SWE-Bench} style tasks},
  url    = {https://huggingface.co/datasets/nvidia/Nemotron-SFT-SWE-v3},
}

@inproceedings{wang2024openhands,
  author    = {Xingyao Wang and Boxuan Li and Yufan Song and Frank F. Xu and others},
  title     = {{OpenHands}: An Open Platform for {AI} Software Developers as Generalist Agents},
  booktitle = {International Conference on Learning Representations (ICLR)},
  year      = {2025},
}

@misc{chen2021codex,
  author       = {Mark Chen and others},
  title        = {Evaluating Large Language Models Trained on Code},
  year         = {2021},
  howpublished = {arXiv:2107.03374},
}

@misc{qwen2026qwen35,
  author = {{Qwen Team}},
  title  = {{Qwen3.5}: Towards Native Multimodal Agents},
  year   = {2026},
  note   = {Checkpoint: \url{https://huggingface.co/Qwen/Qwen3.5-4B}},
  url    = {https://qwen.ai/blog?id=qwen3.5},
}

@misc{qwen2025qwen3,
  author       = {{Qwen Team}},
  title        = {{Qwen3} Technical Report},
  year         = {2025},
  note         = {Checkpoint: \url{https://huggingface.co/Qwen/Qwen3-4B-Thinking-2507}},
  howpublished = {arXiv:2505.09388},
}
